\documentclass[lettersize,journal]{IEEEtran}
\usepackage{amsmath,amsfonts}
\usepackage{algorithmic}
\usepackage{array}
\usepackage[caption=false,font=normalsize,labelfont=sf,textfont=sf]{subfig}
\usepackage{textcomp}
\usepackage{stfloats}
\usepackage{url}
\usepackage{verbatim}
\usepackage{graphicx}
\def\BibTeX{{\rm B\kern-.05em{\sc i\kern-.025em b}\kern-.08em
    T\kern-.1667em\lower.7ex\hbox{E}\kern-.125emX}}
\usepackage{balance}
\usepackage{subfiles}
\usepackage{booktabs}
\usepackage{multirow}
\usepackage{cite}

\DeclareSubrefFormat{parens}{#1(#2)}

\begin{document}
\title{DALNet: A Rail Detection Network Based on Dynamic Anchor Line}
\author{Zichen~Yu,
        Quanli~Liu,~\IEEEmembership{Member,~IEEE},
        Wei~Wang,~\IEEEmembership{Senior Member,~IEEE}, \\
        Liyong~Zhang,~\IEEEmembership{Member,~IEEE},
        and Xiaoguang~Zhao
        \IEEEcompsocitemizethanks{
            \textit{(Corresponding author: Quanli Liu).} 

            \IEEEcompsocthanksitem~Zichen Yu, Quanli Liu, Wei Wang and Liyong Zhang are with the School of Control Science and Engineering, 
            Dalian University of Technology, Dalian 116024, China, and also with the Dalian Rail Transmit Intelligent Control and 
            Intelligent Operation Technology Innovation Center, Dalian 116024, China (e-mail: yuzichen@mail.dlut.edu.cn; liuql@dlut.edu.cn; 
            wangwei@dlut.edu.cn; zhly@dlut.edu.cn).
            \IEEEcompsocthanksitem~Xiaoguang Zhao is with the Dalian Rail Transmit Intelligent Control and Intelligent Operation Technology 
            Innovation Center, Dalian 116024, China, and also with the Dalian Seasky Automation Co., Ltd, Dalian 116024, China (e-mail: xiaoguang.zhao@dlssa.com).
        }
}

\maketitle

\begin{abstract}
    Rail detection is one of the key factors for intelligent train.
    In the paper, motivated by the anchor line-based lane detection methods, we propose a rail detection network 
    called DALNet based on dynamic anchor line. Aiming to solve the problem that the predefined anchor line is image 
    agnostic, we design a novel dynamic anchor line mechanism. It utilizes a dynamic anchor line generator to dynamically 
    generate an appropriate anchor line for each rail instance based on the position and shape of the rails in the input image.
    These dynamically generated anchor lines can be considered as better position references to accurately localize the rails than the 
    predefined anchor lines. In addition, we present a challenging urban rail detection dataset DL-Rail with high-quality annotations and scenario diversity. 
    DL-Rail contains 7000 pairs of images and annotations along with scene tags, and it is expected to encourage the development of 
    rail detection. We extensively compare DALNet with many competitive lane methods. The results show that our DALNet achieves state-of-the-art 
    performance on our DL-Rail rail detection dataset and the popular Tusimple and LLAMAS lane detection benchmarks. 
    The code will be released at \url{https://github.com/Yzichen/mmLaneDet}.
\end{abstract}

\section{Introduction}
\label{sec:intro}
\IEEEPARstart{I}{ntelligent} train uses advanced sensors and algorithms to assist or replace the driver in 
perceiving and analysing the train's operating environment, which can reduce the risk of accidents caused by 
human drivers, and has gradually become a mainstream trend in the development of rail transit. As an important 
safety guarantee for train operation, active obstacle detection has become one of the necessary functions of 
intelligent train~\cite{guan2022, ye2022}. In particular, the accurate rail detection can provide the travelling 
limits of the train as an important prerequisite for active obstacle detection, so it should be regarded as an 
important research area like lane detection.

Despite the importance of the rail detection task, its development is relatively slow. One important reason is 
that there is still a lack of public challenging rail detection datasets. Therefore, we firstly mount cameras on 
a 202-line tram in Dalian, Liaoning Province, China, to collect data, and creat the rail detection dataset DL-Rail, 
which consists of 7000 pairs of images and annotations along with scene tags. Most of the images in the existing 
rail detection datasets~\cite{wang2019railnet, li2022rail} are collected in intercity railway scenarios, which are 
relatively homogeneous. Compared with them, our dataset is collected under urban railway scenarios, which contains 
richer and more complex railway scenarios under different rights-of-way, diverse  weather, and varying  lighting 
conditions. For example, in public right-of-way scenarios, the oncoming vehicles and pedestrians may cause occlusion 
of the rails, which is a situation that is missing in existing datasets.

\begin{figure}[t]%
  \centering
  \subfloat[predefined anchor lines]{
      \includegraphics[width=0.8\linewidth]{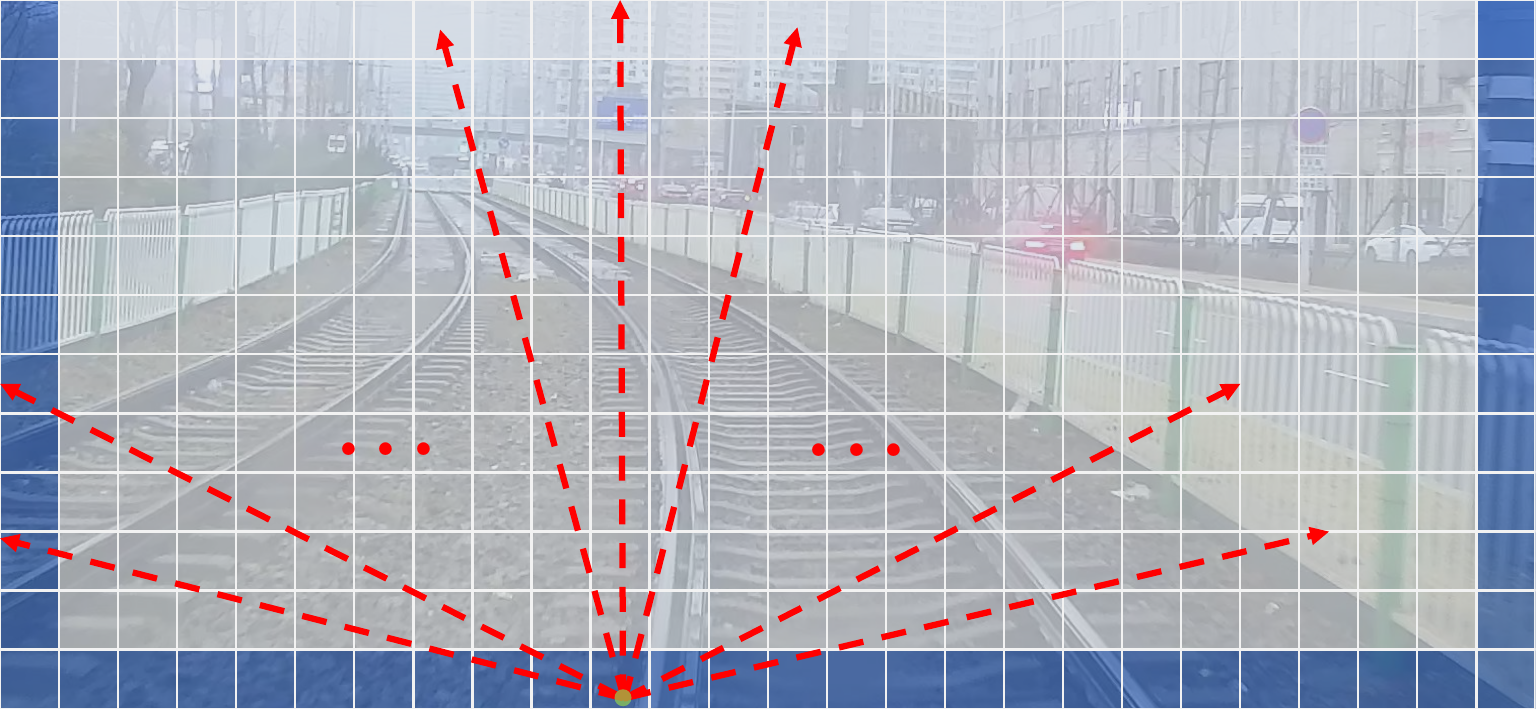}
      \label{fig:intro_predefined}
      }\\
  \subfloat[dynamic anchor lines]{
      \includegraphics[width=0.8\linewidth]{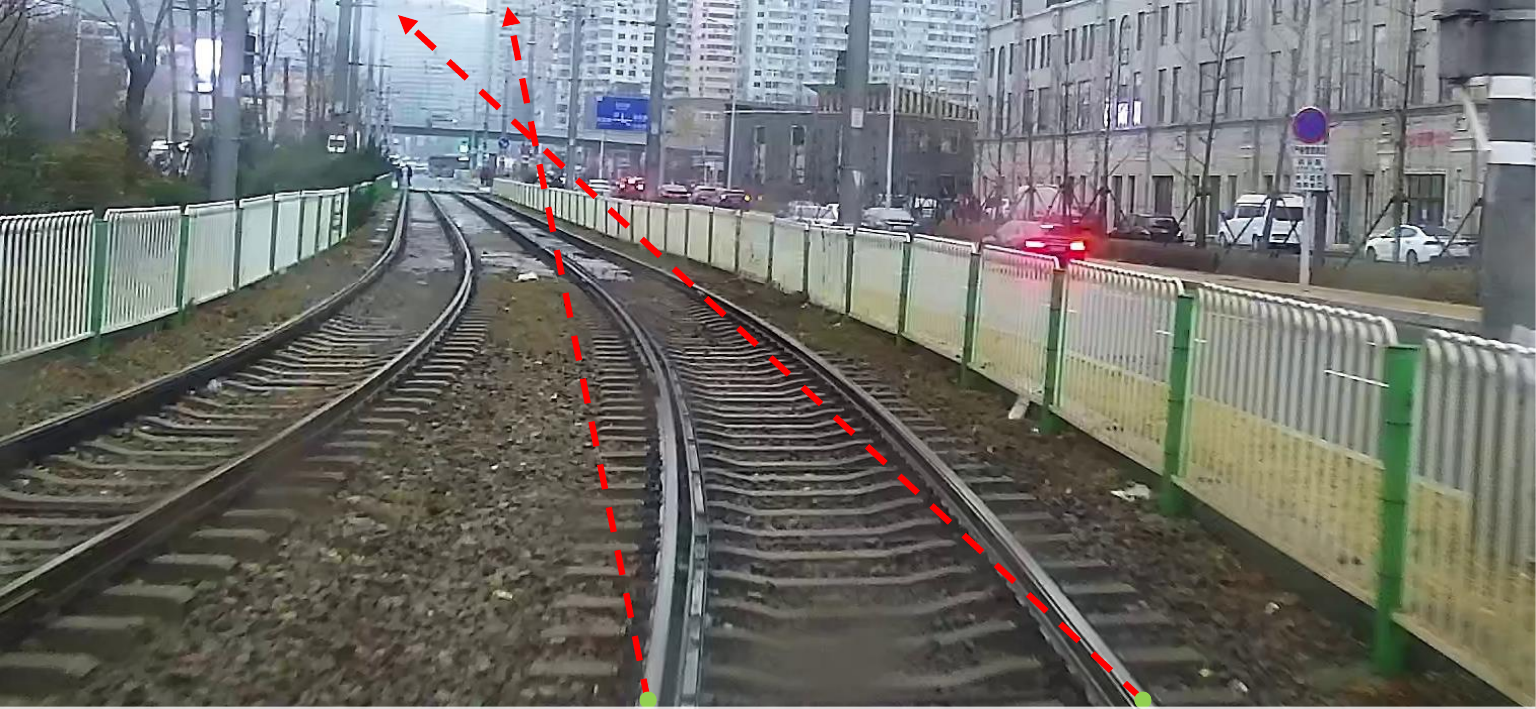}
      \label{fig:intro_dynamic}
      }\\    
  \caption{Illustration of different anchor lines.
  (a) The predefined anchor lines are generated by placing fixed anchor lines with different slopes at each image boundary point. 
  The blue grid cells represent the image boundary regions and the red dashed arrows represent the corresponding anchor lines.
  (b) The dynamic anchor lines are generated based on the position and shape of the rails in the input image, they correspond to the rails one by 
  one and fit more closely to the rails.
  }
  \label{fig:1}
\end{figure}

With this dataset in hand, we turn our attention to how to design a good rail detection algorithm. Previous rail detection 
methods can be roughly classified into traditional hand-crafted rail detection methods and deep learning-based rail detection 
methods. Among them, traditional hand-crafted rail detection methods~\cite{2012nassu, qi2013efficient, zwemer2015vision, zheng2021rail} 
usually use edge features and straight-line features of the rail for detection, but they are not robust enough to cope with 
real-world environments. Deep learning based rail detection methods~\cite{wang2019railnet, li2022rail, Wang2021fusion} use Convolutional 
Neural Networks (CNNs) to extract features and are more advantageous in terms of accuracy and robustness. More importantly, many well-developed 
deep learning based lane detection methods~\cite{qin2020ultra, laneATT, liu2021condlanenet, wang2022ganet} can also provide many valuable 
references for their design, since lanes are similar to rails in that they all have slender shapes. 

The lane detection methods can be classified into segmentation-based methods~\cite{pan2018SCNN, zheng2021resa, pang2022fast_HBNet, neven2018lanedet, sun2023ris}, 
anchor-based methods~\cite{qin2022ultra, liu2021condlanenet, li2019line, laneATT, CLRNet}, keypoint-based methods~\cite{ko202pinet, qu2021folo, wang2022ganet, yao2022devnet},
and curve parameter-based methods~\cite{tab2021polylanenet, liu2021lstr, wang2020polynomial, 2022fengbezier, chen2023bsnet}. In particular, 
anchor-based methods are well suited for rail detection due to their effectiveness and high efficiency. 
Specifically, LaneATT~\cite{laneATT} places a large number of anchor lines (\emph{e.g.}, 2782) at the image boundaries as references 
as illustrated in Fig.~\ref{fig:1}(a), and then uses feature pooling to obtain features of the region of interest (RoI) corresponding 
to the anchor lines, which are used to predict the offsets between the lanes and the anchor lines. When applied to rail detection, 
these predefined anchor lines are able to naturally distinguish between different rail instances and also mitigate the occlusion 
problem by predicting the rail as a whole unit. 
However, this also introduces several problems. Firstly, the number of anchor lines is excessive for rails, and only very few anchor 
lines are actually active during inference, which results in a waste of computational and memory resources. And simply reducing the 
number of anchor lines will cause them to fail to cover the potential locations of the rails. Secondly, these anchor lines are 
image-agnostic and may have a large deviation from the rails, posing difficulties in the localization and identification of the rail. 
In addition, these anchor lines lead to many duplicate predictions that require the non-maximum suppression (NMS) post-processing to 
eliminate, introducing additional time-consumption and complexity. 

To overcome the above limitations, in this paper, we design a dynamic anchor line mechanism, based on which a novel rail detection model 
called DALNet is proposed. As shown in the Fig.~\ref{fig:1}(b), it dynamically generates an appropriate anchor line for each rail instance 
according to the position and shape of the rail in the input image instead of predefined lines. 
Specifically, we propose a dynamic anchor line generator consisting of three parallel lightweight prediction branchs that predict the 
starting point heatmap, the starting point compensation offset and the slope (which represents the angle between the anchor line and the 
image x-axis) of the anchor line, respectively. The corresponding anchor line for each rail instance can then be constructed from the predicted 
starting point coordinate and slope. In the training phase, we utilize the starting point of the rail to generate the starting point label of the 
anchor line, and compute the average angle between the sampled points on the rail and the starting point as the ground truth value of the slope, 
so that the dynamically generated anchor line is more fitting to the rail, and can be used as a better reference. In addition, in the inference phase, 
in order to construct the anchor line, we use a simple max pooling and AND operation to find the peaks in the starting point heatmap, which can eliminate 
duplicate predictions and is more efficient than NMS.

To summarise, our main contributions are summarized as follows::
\begin{itemize}

\item We propose a novel dynamic anchor line-based rail detection model called DALNet, where a dynamic anchor line generator generates an appropriate anchor 
line for each rail instance based on the input image, which not only provide a better reference for rail detection, but also eliminates the need for 
time-consuming NMS post-processing.

\item We present an urban rail detection dataset DL-Rail with high-quality annotations and scenario diversity, expected to facilitate the development of 
rail detection.

\item Without bells and whistles, our proposed DALNet achieves state-of-the-art performance on our DL-Rail dataset. Moreover, it has high accuracy while 
ensuring high efficiency, \emph{e.g.}, a 96.43 F1@50 score and 212 FPS on DL-Rail.

\item Inspiringly, DALNet also shows superior performance on the popular lane detection benchmarks Tusimple~\cite{tusimple} and LLAMAS~\cite{llamas}, which further 
proves the effectiveness of our DALNet and also shows that the dynamic anchor line mechanism also has application potential in the field of lane detection.
\end{itemize}

\section{Related Works}
\label{sec:related-work}

\subsection{Rail Detection Methods}
According to the manner of feature extraction, rail detection methods can be categorized into image 
processing-based methods and deep learning-based methods.

\subsubsection{Image processing-based methods}
Image processing-based methods usually utilize hand-crafted geometric features of the rails to detect the rails. 
Nassu \emph{et al.}~\cite{2012nassu} extract the rails by matching the edge features with the rail patterns modeled as 
a sequence of parabolic segments, where the rail patterns in the near and far regions are precomputed offline and 
generated on-the-fly, respectively.  Qi \emph{et al.}~\cite{qi2013efficient} extract the histogram of oriented gradients~(HOG)
features and generate an integral image, and then detect the rails based on a region growing algorithm. 
Zwemer \emph{et al.}~\cite{zwemer2015vision} propose a generalized tram rail detection system that combines the inverse
perspective transform and a priori geometric knowledge of the rail to find rail segments, and also establishes a 
rail reconstruction algorithm based on graph theory. Zheng \emph{et al.}~\cite{zheng2021rail} segment the input image equally 
along the vertical axis, then approximate the edges as multiple line segments, and finally, fit the rail curve using 
the least squares method after filtering the terminals of the line segments.
These methods are fast enough and do not require a high-performance computing platform, but their performance is 
vulnerable to environmental changes.

\subsubsection{Deep learning-based methods}
Deep learning-based rail detection methods generally rely on CNN for feature extraction. 
Wang \emph{et al.}~\cite{wang2019railnet} convert rail detection into a single-category semantic segmentation task and 
utilize the top-down pyramid structure to extract features, showing good results on their self-built dataset RSDS. 
Wang \emph{et al.}~\cite{Wang2021fusion} also adopt the semantic segmentation scheme and propose a multi-scale prediction network 
to predict the rail area and the forward train area, which makes better use of global information and local characteristics 
by adequately fusing high-resolution and low-resolution feature maps. 
Motivated by the lane detection method UFLD~\cite{qin2020ultra}, Li \emph{et al.}~\cite{li2022rail} propose an efficient 
row-based rail detection method, which first samples the row anchors uniformly along the vertical direction of the image,
and then localizes the rails by predicting the categorical position distributions corresponding to each row anchor.
While these deep learning-based methods are more robust and accurate than image processing-based methods, they are relatively 
slow to develop and rely more on segmentation methods with unsatisfactory performance.

\subsection{Lane Detection Methods}
According to the representation of the lane, existing CNN-based lane detection can be divided into four categories: 
segmentation-based methods, anchor-based methods, curve parameter-based methods, and keypoint-based methods.

\subsubsection{Segmentation-based methods}
The segmentation-based lane detection methods adopt a pixel-level prediction formulation to treat lane detection as a 
semantic segmentation or instance segmentation problem.
Since conventional CNNs are ineffective in detecting objects with slender structures or being occluded, 
SCNN~\cite{pan2018SCNN} proposes a slice-by-slice CNN structure to pass messages between adjacent rows or columns in the 
feature maps, which can extract richer spatial information but is very time-consuming.
RESA~\cite{zheng2021resa} enables horizontal or vertical information aggregation in a parallel manner by recurrently 
shifting sliced feature maps. It not only reduces the time cost, but also prevents information loss during propagation.
Fast-HBNet~\cite{pang2022fast_HBNet} uses the composite transform and the proposed hybrid branching network to generate 
four feature maps with different receptive fields and spatial contexts, allowing the model to capture lane features of 
different shapes, scales, and views. These above methods assign a unique semantic category to different lane lines and require 
a predefined number of lanes. 
To cope with the problem of lane changes, Lanedet~\cite{neven2018lanedet} treats lane detection as an instance segmentation 
problem, which disentangles the binary segmentation results into different lane instances using the predicted lane embeddings.
\cite{sun2023ris} additionally introduces a lane center prediction branch and proposes a center-based lane discrimination method 
instead of the commonly used time-consuming Mean-shift clustering~\cite{2002Comaniciu}, which improves the efficiency and robustness.

\subsubsection{Anchor-based methods}
Anchor-based lane detection methods can be divided into row anchor-based methods and line anchor-based methods. 

The row anchor-based methods formulate lane detection as a classification problem on predefined row anchors uniformly 
distributed over the image. 
UFLD~\cite{qin2020ultra} is a pioneering work using this scheme, which uses global features to directly predict lane position at 
each row anchor and achieves ultra-fast inference. 
Based on this, UFLDV2~\cite{qin2022ultra} proposes a hybrid anchors scheme to alleviate the magnified localization error problem. 
CondLaneNet~\cite{liu2021condlanenet} proposes a conditional lane detection strategy based on conditional convolution and 
row anchor formulation, which solves the lane instance level discrimination problem with lane starting points detection.

For the line anchor-based methods, they lay many anchor lines on the image as references and accurately localize the lanes by 
regressing the offsets.
Inspired by the anchor box in object detection field, Line-CNN~\cite{li2019line} is the first to propose to use the rays with 
different angles emitted from the image boundaries as the anchor lines, and then to predict the offset between the anchor line 
and the lane at each sliding window position over three boundaries.
Following the same anchor lines setting, LaneATT~\cite{laneATT} further introduces anchor-based feature aggregation and attention 
mechanisms to explore the local and global information of the lane, respectively, resulting in a better accuracy-speed tradeoff.
CLRNet~\cite{CLRNet} uses learnable sparse anchor lines which will be optimized as network parameters during training. The 
optimized anchor lines represent the potential locations of lanes in the dataset. In conjunction with a cross-level lane refinement 
scheme, it achieves superior lane detection results. 
However, both predefined and learnable anchor lines are fixed and image-agnostic. In this paper, we propose a novel dynamic anchor 
line mechanism to dynamic generate an appropriate anchor line for each rail instance as a better reference.

\subsubsection{Curve parameter-based methods}
The curve parameter-based lane detection methods fit the lane by predicting the curve parameters. 
PolyLaneNet~\cite{tab2021polylanenet}, LSTR~\cite{liu2021lstr}, and PRNet~\cite{wang2020polynomial} model the lane as a 
polynomial curve (e.g., $x=ay^3+by^2+cy+d$), and the polynomial coefficients corresponding to each lane instance are predicted. 
However, the learning of the abstract polynomial coefficients is not trivial, causing their performance to lag behind other methods.
B$\acute{e}$zierLaneNet~\cite{2022fengbezier} and BSNet~\cite{chen2023bsnet} respectively utilize the parametric B$\acute{e}$zier 
curve and B-spline curve to represent a lane. They transform lane detection into a control points prediction task and a
chieve results competitive with other kinds of methods.

\subsubsection{Keypoint-based methods}
The keypoint-based lane detection methods transform the lane detection task into a keypoint detection and association 
problem. 
PINet~\cite{ko202pinet} predicts keypoints and feature embeddings by stacked hourglass modules~\cite{Newell2016StackedHN}, and 
clusters keypoints based on the similarity of feature embeddings during post-processing. 
FoLoLane~\cite{qu2021folo} constructs the local geometry by predicting the offset from each pixel to its three neighboring 
keypoints, which naturally enables the local association between keypoints. 
GANet~\cite{wang2022ganet} and DevNet~\cite{yao2022devnet} propose a more efficient global association strategy that 
finds the corresponding lane instance by predicting the offset from each keypoint to the lane starting point.

\begin{figure*}[ht]
    \centering
    \includegraphics[width=0.95\linewidth]{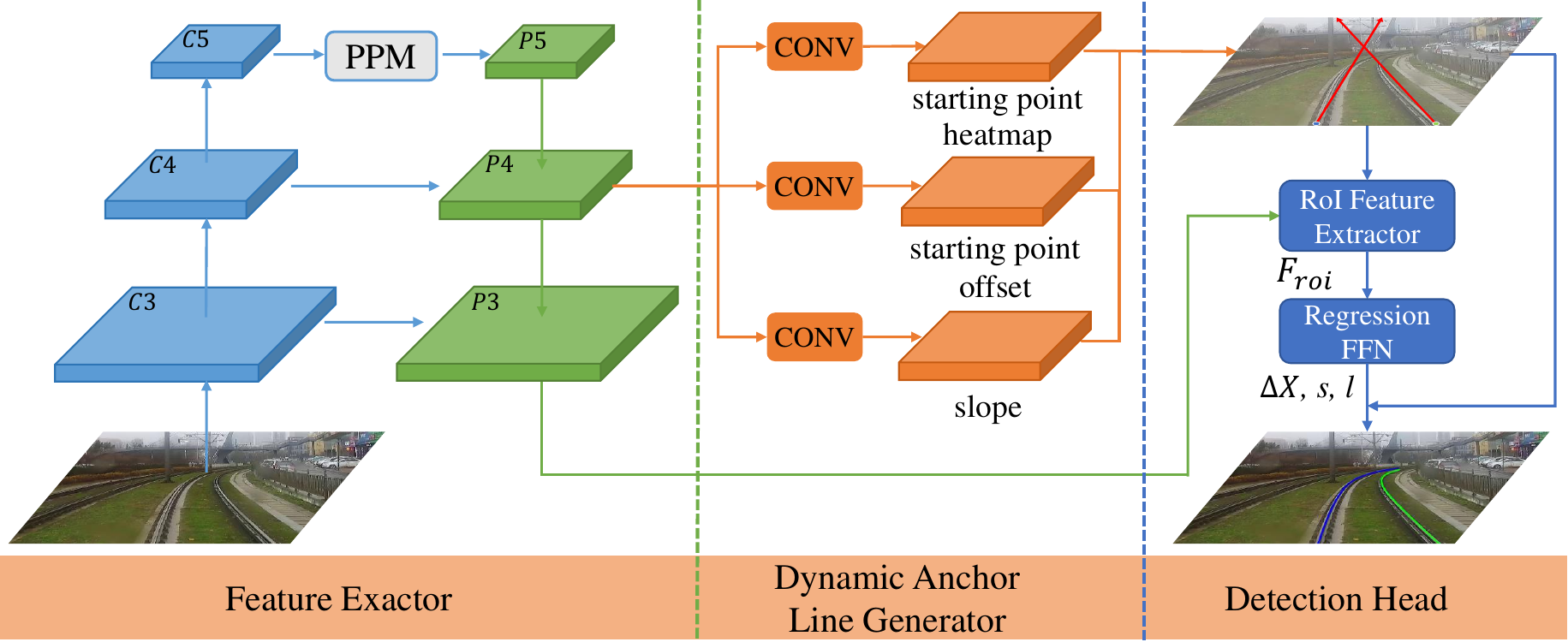}
    \caption{Framework of DALNet. DALNet consists of three main components: feature extractor, dynamic anchor line generator
    and rail detection head. Given an input image, the feature extractor extracts multi-scale features and the pyramid pooling 
    module (PPM) is employed to enlarge the receptive field. The dynamic anchor line generator includes three parallel branches 
    that predict the starting point, offset and slope of the anchor line. Based on these predictions, the generator constructs 
    a fitting anchor line for each rail instance. In the detection head, the generated anchor line is taken as a reference 
    to localize the rail by predicting the horizontal offset between the rail and the anchor line.}
    \label{fig:arch}
\end{figure*}

\section{Method}
\label{sec:method}
The essence of our DALNet framework is to dynamically generate an appropriate anchor line as reference for each rail 
instance based on the position and shape of the rails in the image, instead of fixed and image-agnostic anchor lines. 
The overall architecture is shown in Fig.~\ref{fig:arch}.

The DALNet is a simple and efficient network consisting of a feature extractor, a dynamic anchor line generator and a rail 
detection head. 
Given a forward-looking image $ I\in \mathbb{R} ^{3 \times H_{I} \times W_{I}} $ taken by a camera mounted in front of the 
train, this image is fed to the feature extractor to generate multi-scale rich features, where a pyramid pooling module (PPM)~\cite{zhao2017pyramid} 
is used to enlarge the receptive field and gather global context information.
In the dynamic anchor line generator, the image features are fed into three parallel branches to predict the starting point, 
offset, and the slope of the anchor line, then an anchor line is dynamic constructed for each rail instance based on these 
predicted results. This generated anchor line is used as a reference in the rail detection head to accurately localize the 
rail by predicting the horizontal offset.

\subsection{Rail and Anchor Representation}
Rails are similar to lanes in that they both have a slender appearance structure. Therefore, we follow the lane representation 
in~\cite{laneATT,li2019line} by representing a rail as a sequence of $N_{pts}$ points, \emph{i.e.}, $P=\{(x_0, y_0), (x_1, y_1), \cdots, (x_{N_{pts}-1}, y_{N_{pts}-1}) \}$, 
where the $y$ coordinates of these points are fixed and uniformly sampled along the vertical axis of the image, \emph{i.e.}, $y_{i} =  \frac{H_{I}}{N_{pts}-1}\times{i}$. 
In addition, the start index $s$ and the length $l$ are used to describe the range of the rail in the vertical axis.

Different from comman object detection~\cite{ren2017fast, lin2020retinanet}, we utilize anchor lines instead of anchor boxes to 
provide the position priors for rails, since rails are usually global and slender. The anchor line can be regarded as a ray emitted 
from the image boundary with a certain slope, so it can be parameterized as $(x_{start}, y_{start}, \theta)$, where $(x_{start}, y_{start})$ 
denotes the starting point coordinate and $\theta$ denotes the slope.
However, existing methods usually use predefined or learnable anchor line parameters, which are independent of the current input image.
In this paper, we use a new dynamic anchor line mechanism. It determines the anchor line parameters according to the position and shape of 
the rails in the input image and dynamically constructs a fitting anchor line for each rail instance.

\begin{figure}[t]
    \centering
    \includegraphics[width=0.99\linewidth]{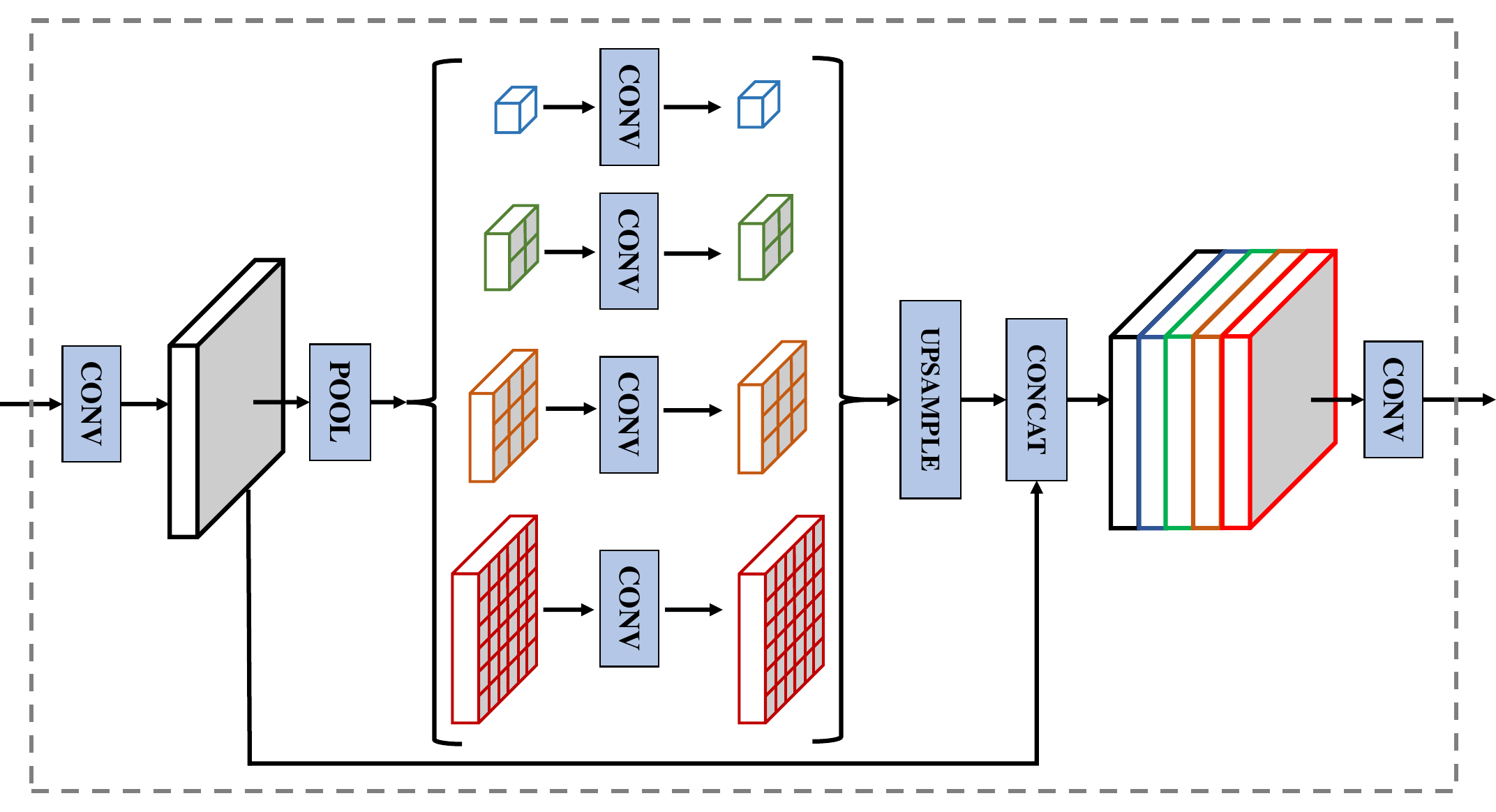}
    \caption{
        The structure of the PPM. PPM performs pooling operations with different scales on the input feature map to establish 
        a pooled feature pyramid. By fusing features at different pyramid scales through concatenation operation, PPM effectively 
        enlarge the receptive field and gathers global context information.
    }
    \label{fig:ppm}
\end{figure}

\subsection{Feature Exactor}
Since both high-level semantic features and low-level geometric features are important for the rail detection, a feature pyramid 
network (FPN)~\cite{lin2017feature} based on the standard ResNet architecture (\emph{e.g.}, ResNet18) is employed to combine the 
multi-level features of the input image by constructing a feature pyramid with levels $P3$ to $P5$. To ensure efficiency, all pyramid 
feature maps have $C=64$ channels.

In addition, the generation of dynamic anchor line requires the perception of the overall characteristics of the rail, so the ability 
to establish long-distance dependencies is necessary. To this end, we introduce the PPM~\cite{zhao2017pyramid} which has been shown 
to be effective in semantic segmentation, and insert it between the ResNet backbone and the FPN neck. As shown in Fig.~\ref{fig:ppm}, 
The PPM pools the input feature map to different sizes, creating a pyramid of pooled features, with each layer of the pyramid having a 
different receptive field. By upsampling the pooled features and then concatenating them together for fusion, it can effectively enlarge 
the receptive field and capture global information at different scales.
In the current work, we set up four fixed-size average pooling modules for PPM:~$1\times1$, $2\times2$, $3\times3$ and $6\times6$.

\subsection{Dynamic Anchor Line Generator}
The purpose of the dynamic anchor line generator is to generate a fitting anchor line for each rail instance based on the extracted 
image features. The anchor line can be determined based on its starting point and slope, thus in this paper, we decompose the generation 
of the anchor line into the starting point estimation and the slope estimation tasks.

\subsubsection{Starting Point Heatmap Estimation}
To estimate the starting point location, we design a keypoint detection branch consisting of two convolutional layers. It takes the $P4$ 
feature map as input and outputs a heatmap $ \hat{P} \in \mathbb{R}^{H_{4} \times W_4} $ to represent the probability that each pixel belongs 
to the starting point.

In the training stage, we simply use the starting point of the rail to generate the ground-truth heatmap. Specifically, Given a set of starting 
points of the rails $ p_k=(x_k, y_k), k=1, \cdots ,N $, their downsampled positions $ \tilde{p}_k = (\tilde{x}_k, \tilde{y}_k)=(\lfloor x_k/\delta \rfloor, \lfloor y_k/\delta \rfloor)  $
in the feature map are computed, where $\delta=16$ is the stride of the feature map. Then the ground-truth heatmap $P$ is calculated as a smooth 
gaussian region of width $\sigma$ around the starting point. The label at position $(i, j)$ of the resulting heatmap $P$ is given by:
\begin{equation}
    P(i, j)= \max\limits_{k} \exp(- \frac{(i-\tilde{x}_k)^2 + (j-\tilde{y}_k)^2}{2\sigma^2}),
\end{equation}
where $(i, j) \in[0, W_4]\times[0,H_4]$ is the spatial indices of the heatmap and the standard deviation $\sigma$ is related to the size of 
the input image.

The penalty-reduced focal loss~\cite{law2018cornernet} is applied to the predicted heatmap to resolve the imbalance between the starting point region and the 
non-starting point region. Let $\hat{P}_{ij}$ and $P_{i,j}$ be the predicted and ground-truth label at the location $(i, j)$ in the 
heatmap respectively, the starting point estimation loss is defined as:
\begin{equation}
    \mathcal{L}_{heat} = - \frac{1}{N_p} \sum\limits_{j=1}^{H_4}\sum\limits_{i=1}^{W_4} 
    \begin{cases}
        {(1-\hat{P}_{ij})^{\alpha}log(\hat{P}_{ij})} & P_{ij} = 1\\
        (1-P_{ij})^{\beta}\hat{P}_{ij}^{\alpha}log(1-\hat{P}_{ij}) & P_{ij} \neq 1
    \end{cases},
\end{equation}
where $\alpha$ and $\beta$ are adjustable hyperparameters, $N_p$ is the number of rails in the image, and the 
term $(1-P_{ij})^\beta$ is used to reduce the penalty around the ground-truth position.

\subsubsection{Starting Point Offset Estimation}
The peak of the predicted heatmap gives a coarse position estimate of the starting point accurate to the stride $\delta$ of the feature map. 
In order to localize the starting point more precisely, we append an additional offset estimation branch to predict an offset map $ \hat{O} \in \mathbb{R}^{H_{4} \times W_4 \times 2} $.
The predicted offset map has two main functions. On the one hand, it eliminates the quantization error introduced during the downsampling process, 
in which we map the starting point $p$ to the downsampled position $\tilde{p}$; on the other hand, it can correct the starting point position 
when the starting point predicted by the above heatmap is biased.

In the training stage, we select a square region within distance $r$ around the ground-truth starting point as the valid region and only 
apply a smooth L1 loss~\cite{ren2017fast} on this valid region to constrain the predicted offset map, as follows:
\begin{equation}
    \begin{aligned}
        \mathcal{L}_{offset}= &\frac{1}{N_p*(2r+1)^2} \sum\limits_{p} \sum\limits_{t_y=-r}^{r} \sum\limits_{t_x=-r}^{r} SmoothL1Loss(\\
        &\hat{O}_{\tilde{p}+(t_x, t_y)}, \frac{p}{\delta}-\tilde{p}-(t_x, t_y)),
    \end{aligned}
\end{equation}
where $t_x$ and $t_y$ are the offsets of the valid position relative to $\tilde{p}$ along the x-axis and y-axis.

\subsubsection{Slope Estimation}
A slope estimation branch is added to predict a slope map $\hat{\Theta} \in \mathbb{R}^{H_{4} \times W_4}$, which determines the slope 
of the generated anchor line. A proper slope will make the anchor line fit the rail more closely, making the subsequent rail detection based 
on the anchor line easier and more precise.
Given a rail $L=\{(x_s, y_s), (x_{s+1}, y_{s+1}), \cdots, (x_e, y_e)\}$ (here only the valid points of the rail are given for brevity), where 
$e=s+l-1$ is the cutoff index of the rail. We compute the average slope of all lines connecting the non-starting point and the starting 
point on the rail as the ground-truth slope as follows:

\begin{equation}
    \theta= \frac{1}{l} \sum\limits_{i=s}^e arctan2( \mid y_i-y_s \mid , x_i-x_s),
\end{equation}

In the training phase, the same valid training region is chosen as the offset estimation and the L1 loss is imposed to constrain 
the slope prediction, as follows:
\begin{equation}
    \begin{aligned}
        \mathcal{L}_{slope}= &\frac{1}{N_p*(2r+1)^2} \sum\limits_{k=1}^{N_p} \sum\limits_{t_y=-r}^{r} \sum\limits_{t_x=-r}^{r}\\
        &L1Loss(\hat{\Theta}_{\tilde{p}_k+(t_x, t_y)}, \theta_k),
    \end{aligned}
\end{equation}
where $k$ is the index of the rail.

\subsubsection{Gather Indices and Decode}
In order to decode the corresponding anchor line, we utilize the max pooling and AND operations to find the peaks in the 
predicted starting point heatmap, and then determine the starting point position and slope based on the peak position, 
respectively. The max pooling operation naturally eliminates duplicate predictions and it is more efficient than 
distance-based NMS~\cite{laneATT} and IoU-based NMS~\cite{CLRNet}.

Specifically, after the max pooling and AND operations, we can easily obtain the coordinate $(\hat{x},\hat{y})$ of each peak 
in the starting point heatmap. This coordinate is used as a rough starting point position on the one hand, and to associate 
the corresponding offset and slope on the other.
Thus, the final anchor line is determined as $(\delta(\hat{x}+ \hat{o}_x), \delta( \hat{y}+ \hat{o}_y), \hat{\theta})$, 
where $(\hat{o}_x, \hat{o}_y)$ is the predicted offset at location $(\hat{x},\hat{y})$ in $\hat{O}$, 
and $\hat{\theta}$ is the predicted slope at location $(\hat{x},\hat{y})$ in $\hat{\Theta}$.

\subsection{Rail Detection Head}
In the detection head, the generated anchor line is employed as a reference to accurately localize the final rail.

First, we first extract the RoI feature of the anchor line from the $P3$ feature map. Given the generated anchor 
line $(x_{start}, y_{start}, \theta)$, we sample $N_s$ points uniformly on it, and then utilize 
bilinear interpolation to compute the feature of each sampling point. The features of all sampling points are 
concatenated as RoI features $F_{roi}$. 

Then, to generate the rail proposal, the RoI feature are fed to a fully connected layer, producing $N_{pts}$ horizontal 
offsets $\Delta{X}=\{\Delta{x_0},\Delta{x_1},\cdots,\Delta{x_{N_{pts}-1}}\}$ between the rail and the anchor line, 
the starting index $s$ and the length $l$ of the rail. 
In this way, for each fixed $y_{i} =  \frac{H_{I}}{N_{pts}-1}\times{i}$, 
each corresponding predicted x-coordinate $\hat{x}_i $ is calculated as follows:
\begin{equation}
    \begin{aligned}
        \hat{x}_{i}= \frac{1}{tan\theta} \cdot (y_i-y_{start})+x_{start}+\Delta{x_i}.
    \end{aligned}
\end{equation}

The overall loss is defined as follows:
\begin{equation}
    \begin{aligned}
        \mathcal{L}_{total}= &\lambda_{heat}\mathcal{L}_{heat}+ \lambda_{offset}\mathcal{L}_{offset}+ \lambda_{slope}\mathcal{L}_{slope}\\
        +&\lambda_{sl}\mathcal{L}_{sl}+\lambda_{LIoU}\mathcal{L}_{LIoU}
    \end{aligned}
\end{equation}
where $\mathcal{L}_{sl}$ is the smooth L1 loss to supervise the prediction of starting index $s$ and 
length $l$. $\mathcal{L}_{LIoU}$ is the Line IoU loss~\cite{CLRNet}, which is calculated based on the predicted rail
and the ground-truth rail, and it is used to constrain the offsets regression $\Delta{X}$. 
$\lambda_{heat}$, $\lambda_{offset}$, $\lambda_{slope}$, $\lambda_{sl}$ and $\lambda_{LIoU}$ are 
the weight coefficients corresponding to the losses, respectively.

\section{Experiments}
In order to fully validate the effectiveness of our DALNet, we conduct experiments on our self-built DL-Rail dataset. Tusimple and LLAMAS datasets respectively.  
Considering that rail detection is a prerequisite for train active obstacle detection, its inference should be fast enough even on 
edge devices with limited computational resources, so as to allow sufficient time for the obstacle detection algorithms. Therefore, 
we only select some rapid lane detection models for comparison, which can reason over 200 FPS on our gpu device, promising to be 
fast enough to run on embedded platforms.

\label{sec:exp}
\subsection{Datasets and Evaluation Metric}
\subsubsection{DL-Rail}
\begin{figure}[t]
    \centering
    \includegraphics[width=0.99\linewidth]{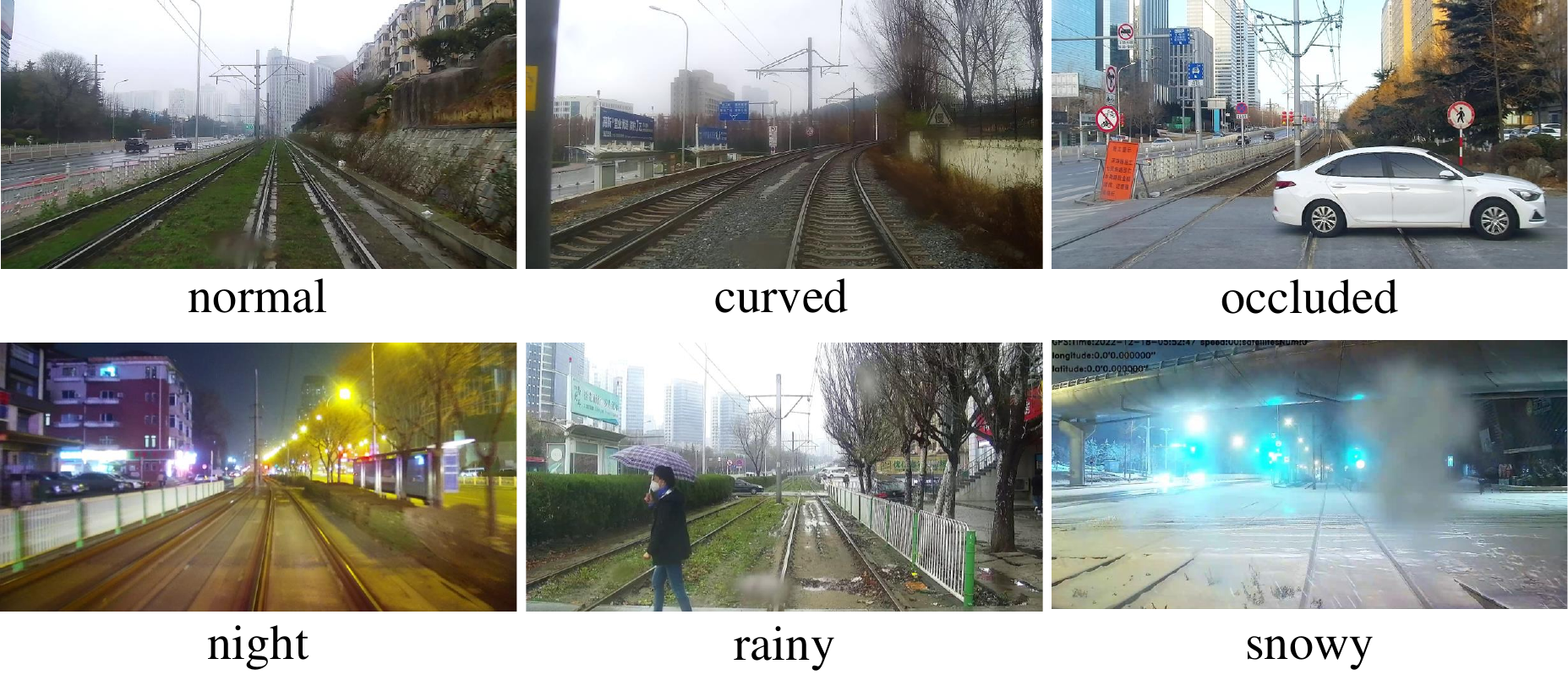}
    \caption{
        Dataset examples for different scenarios in our DL-Rail dataset.
    }
    \label{fig:dataset}
\end{figure}

In this paper, we present a challenging urban rail detection dataset. In order to collect the data, we mount cameras at the 
front of a tram on line 202 in Dalian, Liaoning Province, China, and record videos of forward scenes during the actual train 
operation, where the video resolution is 1920$\times$1080. We carefully handpicked 50 videos covering challenging scenarios such as poor 
lighting and inclement weather. 
In order to prevent duplication or similarity of scenes in adjacent frames, each video is sampled at 2-second intervals and eventually 
7000 images are collected to form the dataset. The dataset is divided into training and test set in the ratio of 8:2. In addition, 
the test set is further divided according to the scenario categories, which facilitates the evaluation of the model's ability to cope 
with various types of challenging scenarios, where the scenario categories include normal, nighty, rainy and so on.
The examples corresponding to these scenarios are given in Fig.~\ref{fig:dataset}.

Compared with other rail detection datasets~\cite{wang2019railnet, li2022rail}, our dataset is more challenging. On the one hand, 
many images in our dataset are captured in mixed right-of-way scenarios where rails are often occluded by oncoming vehicles and pedestrians, 
and on the other hand, many images are recorded under poor lighting and weather conditions where rails are unclear and difficult 
to recognize and localize.

We use the LabelMe tool to manually annotate the ego rails and each rail is represented by a line strip. When 
annotating the rails, the history images and contextual information are combined to accurately label the rails 
if they are occluded or unclear, because the detection of rails in these scenarios is also very important. 
We hope the users of this dataset can design a model that can utilize global contextual or temporal information 
to reason about the location of invisible rails as humans do, instead of just giving up on the detection in such 
scenarios.

The F1-measure is adopted as the metric and it is based on the Intersection-over-union (IoU) between the predicted and ground truth rails. 
To calculate the IoU, we connect both the predicted and ground truth rail points with width 30 to generate the corresponding mask,
and then calculate the IoU between the predicted mask and the ground truth mask. Given an IoU threshold, the prediction is considered 
as true positive when the IoU of the predicted rail is greater than this threshold. With this condition, we separately 
statistic the number of true postive (TP), false postive (FP) and false negative (FN) and calculate the F1-measure as follows:

\begin{equation}
  F1= \frac{Precision*Recall}{Precision+Recall},
\end{equation}
\begin{equation}
  Precision= \frac{TP}{TP+FP},
\end{equation}
\begin{equation}
  Recall= \frac{TP}{TP+FN}.
\end{equation}

In this paper, we report the F1@50 and F1@75 metrics for IoU thresholds of 0.5 and 0.75, respectively. In addition, to 
better compare the performance of the models, we also report a more comprehensive metric mF1 as follows:
\begin{equation}
mF1=(F1@50+F1@55+ \cdots +F1@95)/10.
\end{equation}

\subsubsection{Tusimple}
Tusimple~\cite{tusimple} is a lane detection dataset for real highway scenarios. It contains 3626 training images, 358 validation images 
and 2782 test images, and all images have 1280$\times$720 pixels. The main evaluation metrics for this dataset are false discovery 
rate (FDR), false negative rate (FNR) and accuracy, defined as follows:
\begin{equation}
  FDR= \frac{FP}{TP+FP},
\end{equation}
\begin{equation}
  FNR= \frac{FN}{TP+FN},
\end{equation}
\begin{equation}
  Accuracy = \frac{\sum_{clip}C_{clip}}{\sum_{clip}S_{clip}},
\end{equation}
where $C_{clip}$ is the number of correctly predicted lane points and $S_{clip}$ is the number of all ground turth points in 
the image. For a predicted lane point, it is considered correct only if it lies within 20 pixels of a ground truth point. 
For a predicted lane, it is considered a true positive when its corresponding prediction accuracy is greater than 85$\%$.

\subsubsection{LLAMAS}
LLAMAS~\cite{llamas} is a large-scale lane detection dataset containing over 100,000 images of size 1276$\times$717, of which 58,269 images are 
used for training, 20,844 images for validation and 20,929 images for testing. Unlike other datasets, the lane annotations 
for the LLAMAS dataset are generated automatically with the help of high definition maps. 
Since the annotations of the test set are not public, the inference results are uploaded to the LLAMAS benchmark's website for 
evaluation. The evaluation results include precision, recall and F1-measure.

\subsection{Implementation Details}
All the input images are cropped and resized to 800$\times$320. In order to enhance the generalization, the random horizontal 
flipping and random affine transformations (rotation, translation, scaling) are adopted as data augmentations. 
We set the number of sampled points as $N_s=36$ and $N_{pts}=72$, and the loss weights are set as $\lambda_{heat}=1.0$, 
$\lambda_{offset}=1.0$, $\lambda_{slope}=3.0$, $\lambda_{sl}=0.3$ and $\lambda_{LIoU}=6.0$.  
For optimization, we use the AdamW optimizer with weight decay of 0.01. The learning rate is initialized with $1.0\times10^{-4}$, 
and decayed with a cosine annealing strategy. We train 70 epochs for DL-Rail, 70 epochs for Tusimple, and 20 epochs for LLAMAS. 
All experiments are performed on 4 Nividia 3090 GPUS, with a batchsize of 8 per GPU.

\begin{table*}[t]
    \begin{center}
    \caption{performance on DL-Rail}
    \label{tab:DL-Rail}
    \resizebox{\textwidth}{!}{
        \begin{tabular}{lcccccccccccc}
            \hline
            Method & Backbone & mF1 & F1@50 & F1@75  & FPS & GFlops & Normal & Curved & Occluded & Night & Rainy & Snowy \\ 
            \hline
            \textbf{Curve-based} \\ 
            \hline
            B$\acute{e}$zierLaneNet~\cite{2022fengbezier} & ResNet18 & 42.81 & 85.13 & 38.62  & 244 & \textbf{7.55} & 85.03 & 80.00 & 83.33 & 86.51 & 85.35 & 85.24 \\ 
            \hline
            \textbf{Keypoint-based} \\ 
            \hline
            GANet~\cite{wang2022ganet} & ResNet18 & 57.64 & 95.68 & 62.01 & 208 & 21.27 & 99.00 & 88.00 & 86.67 & 93.17 & 91.22 & 91.07 \\
            \hline
            \textbf{Anchor-based} \\ 
            \hline
            CondLaneNet~\cite{liu2021condlanenet} & ResNet18 & 52.37 & 95.10 & 53.10 & 210 & 10.16 & 97.73 & 90.11 & 86.67 & 93.05 & 91.44 & 91.24 \\
            UFLD~\cite{qin2020ultra} & ResNet18 & 53.50 & 93.67 & 57.74 & \textbf{398} & 8.39 & 95.28 & 86.24 & 85.00 & 92.73 & 91.29 & 90.65 \\
            UFLD~\cite{qin2020ultra} & ResNet34 & 53.76 & 94.78 & 57.15 & 233 & 16.9 & 96.64 & 89.65 & 85.00 & 93.53 & 91.32 & 90.77 \\
            LaneATT(with FPN)~\cite{laneATT} & ResNet18 & 55.57 & 93.82 & 58.97 & 250 & 12.46 & 96.52 & 85.71 & 83.33 & 92.47 & 91.37 & 90.83 \\
            \textbf{DALNet~(Ours)} & ResNet18 & \textbf{59.79} & \textbf{96.43} & \textbf{65.48}  & 212 & 9.75 & \textbf{99.04} & \textbf{92.24} & \textbf{88.33} & \textbf{94.33} & \textbf{93.92} & \textbf{92.11} \\
            \hline
        \end{tabular}
    }
    \end{center}
\end{table*}
\subsection{Results}
\subsubsection{Results on DL-Rail}
As shown in Tab.~\ref{tab:DL-Rail}, we compare the performance on the DL-Rail test set with various fast lane detection methods, 
where the segmentation-based methods are generally slow and therefore not included in the comparison. The results show that our 
proposed DALNet achieves state-of-the-art results on F1@50, F1@75 and mF1 metrics, demonstrating that DALNet has strong capability 
for rail detection. 
In comparison with the curve-based method, DALNet outperforms B$\acute{e}$zierLaneNet~\cite{2022fengbezier} by 16.98 and 11.3 on
mF1 and F1@50. Meanwhile, the poor performance of B$\acute{e}$zierLaneNet~\cite{2022fengbezier} indicates the high freedom of rails 
makes the prediction of control points difficult. 
GANet~\cite{wang2022ganet}, as a keypoint-based method, also shows good performance on the rail detection task. In comparison with it, 
DALNet outperforms it by 2.15 and 0.75 on mF1 and F1@50 with slightly faster inference. 
CondLaneNet~\cite{liu2021condlanenet} and UFLD~\cite{qin2020ultra} are two row-anchor based methods, and DALNet outperforms them 
on F1@50 by 1.33 and 1.65, respectively. However, UFLD has the fastest inference speed among these methods. 
In comparison with the line anchor-based method LaneATT~\cite{laneATT}, DALNet significantly surpasses it by 4.22 and 2.61 on mF1 and F1@50, 
achieving a better trade-off between speed and localization accuracy.

\begin{table}[!htbp]
    \begin{center}
    \caption{performance on tusimple}
    \label{tab:Tusimple}
    \resizebox{0.5\textwidth}{!}{%
        \begin{tabular}{lcccccc}
            \hline
            \textbf{Method} & \textbf{Backbone} & \textbf{Acc (\%)}  & \textbf{FDR (\%)} & \textbf{FNR (\%)}\\ 
            \hline
            \textbf{Curve-based} \\ 
            \hline
            B$\acute{e}$zierLaneNet~\cite{2022fengbezier} & ResNet18 & 95.41  & 5.30 & 4.60 \\ 
            \hline
            \textbf{Keypoint-based} \\ 
            \hline
            GANet~\cite{wang2022ganet} & ResNet18 & 95.95 & \textbf{1.97} & 2.62 \\
            \hline
            \textbf{Anchor-based} \\ 
            \hline
            UFLD~\cite{qin2020ultra} & ResNet18 & 95.82 & 19.05 & 3.92 \\
            UFLD~\cite{qin2020ultra}  &  ResNet34 & 95.86  & 18.91 & 3.75 \\
            LaneATT~\cite{laneATT} & ResNet18 & 95.57 & 3.56 & 3.01 \\
            LaneATT~\cite{laneATT} & ResNet34 & 95.63 & 3.53 & 2.92 \\
            CondLaneNet~\cite{liu2021condlanenet} & ResNet18 & 95.48 & 2.18 & 3.80 \\
            \textbf{DALNet~(Ours)} & ResNet18 & \textbf{96.68} & 2.47 & \textbf{2.50} \\
            \hline
        \end{tabular}
    }
    \end{center}
\end{table}

\subsubsection{Results on Tusimple}
The results on the Tusimple test set are shown in Tab.~\ref{tab:Tusimple}. In comparison with other fast lane detection models, DALNet achieves the best performance 
on both accuracy and FNR metrics. Specifically, compared to the fastest model UFLD~\cite{qin2020ultra}, DALNet has a significantly lower FDR as well as higher localization 
accuracy. Furthermore, The ResNet18 version of our method outperforms LaneATT~\cite{laneATT} (ResNet34) by 1.05 on accuracy, which is already a significant improvement 
for the Tusimple dataset. These comparison results show that our DALNet also has great potential for lane detection tasks.

\begin{table}[!htbp]
    \begin{center}
    \caption{performance on LLAMAS}
    \label{tab:llamas}
    \resizebox{0.5\textwidth}{!}{%
        \begin{tabular}{lcccccc}
            \hline
            \textbf{Method} & \textbf{Backbone} & \textbf{F1 (\%)} &\textbf{Precision (\%)}  & \textbf{Recall (\%)} \\ 
            \hline
            \textbf{Curve-based} \\ 
            \hline
            B$\acute{e}$zierLaneNet~\cite{2022fengbezier} & ResNet18 & 94.91 & 95.71  & 94.13  \\ 
            \hline
            \textbf{Anchor-based} \\ 
            \hline
            LaneATT~\cite{laneATT} & ResNet18 & 93.46 & \textbf{96.92} & 90.24  \\
            LaneATT~\cite{laneATT} & ResNet34 & 93.74 & 96.79 & 90.88  \\
            \textbf{DALNet~(Ours)} & ResNet18 & \textbf{96.12} & 96.83 & \textbf{95.42}\\
            \hline
        \end{tabular}
    }
    \end{center}
\end{table}
\subsubsection{Results on LLAMAS}
The results on LLAMAS benchmark are shown in Tab.~\ref{tab:llamas}. Since this benchmark is relatively new, some methods have not submitted their evaluation 
results. Therefore, only B$\acute{e}$zierLaneNet~\cite{2022fengbezier} and LaneATT~\cite{laneATT} methods are incorporated into the comparison. The results show that our DALNet 
achieves the best performance on F1 and recall metrics, surpassing B$\acute{e}$zierLaneNet~\cite{2022fengbezier} and LaneATT~\cite{laneATT} (ResNet34) on F1 
by 1.21 and 2.38, respectively, which is significant improvement.
These evaluations results are also available on the official benchmark's website~\footnote{\url{https://unsupervised-llamas.com/llamas/benchmark_splines}}.

\begin{figure*}[t]
  \centering
  \includegraphics[width=0.95\linewidth]{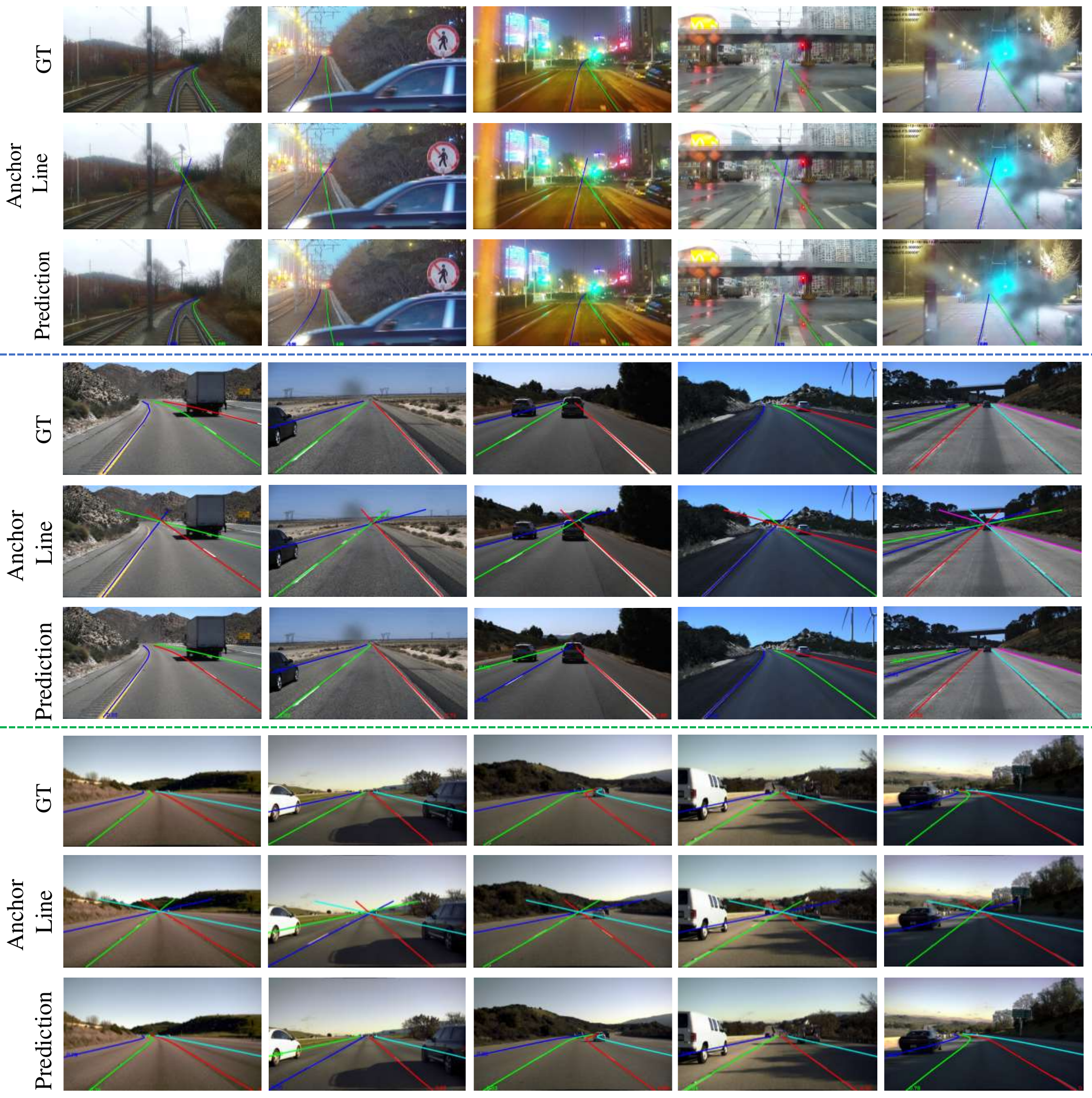}
  \caption{Qualitative results on DL-Rail(the first row), Tusimple(the middle row) and LLAMAS(the last row) datasets.
  Different rail or lane instances are drawn in different colors.} 
  \label{fig:vis_results}
\end{figure*}

\subsection{Ablation Study}
To analyze the effectiveness of each composition in our proposed DALNet, the ablation experiments are conducted on our DL-Rail dataset. 
For our baseline model, we remove the dynamic anchor line generator and directly use fixed anchor lines as references following~\cite{laneATT},\cite{li2019line}. 
To better evaluate our dynamic anchor line mechanism, we further divide it into dynamic anchor line generator (DALG) and dynamic 
anchor line reference (DALR), where DALG is responsible for the dynamic anchor line prediction, and DALR means that the generated dynamic 
anchor line is used as a reference for rail localization in the detection head. Tab.~\ref{tab:ablation} summarizes the ablation results on our DALG, 
DALR and pyramidal pooling modules (PPM).

\begin{table}[!htbp]
    \centering
	\caption{effects of each component in our method}
    \label{tab:ablation}
    \scalebox{1.0}{
        \begin{tabular}{cccc|ccc}
            \hline
            Baseline & DALG & DALR & PPM & F1@50 & F1@75 & mF1 \\ 
			\hline
            \checkmark &  &  &  & 94.68 & 62.81 & 57.91 \\
            \checkmark & \checkmark &  &  & 94.59 & 61.35 & 57.43 \\ 
            \checkmark & \checkmark & \checkmark &  & 96.01 & 64.09 & 59.37 \\ 
			\checkmark & \checkmark & \checkmark & \checkmark & 96.43 & 65.48 & 59.79 \\ 
			\hline
        \end{tabular}
    }
\end{table}

\subsubsection{Effect of dynamic anchor line generator}
As shown in the second row of Tab.~\ref{tab:ablation}, the DALG module is added to the baseline model and is only employed as an auxiliary branch 
to predict the starting point and slope of the anchor line. However, the results on F1@50, F1@75 and mF1 are consistently 
decreased compared to baseline. This indicates that the starting point and slope prediction tasks negatively affect the learning of 
the rail proposal prediction task, which is widely known as ``negative transfer''.

\subsubsection{Effect of dynamic anchor line reference}
As shown in the third row of Tab.~\ref{tab:ablation}, the DALG and DALR are combined to form a complete dynamic anchor line mechanism, 
where the anchor line generated by the DALG module is used as a reference. Comparing the results in the second and third rows, 
we can see that the introduction of DALR improves F1@50, F1@75 and mF1 by 1.42, 2.74 and 1.94, respectively, which indicates that 
the dynamic anchor line reference can effectively improve the localization performance and it is the key to the whole mechanism. 
However, it is worth noting that DALG is a prerequisite for DALR, even though its use alone has negative effects.

\subsubsection{Effect of pyramid pooling module}
To further enlarge the receptive field, the PPM is inserted between the ResNet backbone and FPN neck. It aggregates contexts with 
different scales through pyramid pooling. Comparing the third and fourth rows of Tab.~\ref{tab:ablation}, the 
introduction of PPM improves the mF1 and F1@50 metrics by 0.42, indicating that the aggregation of global context 
can further improve the performance of rail localization.
Furthermore, we compare different ways of global context aggregation in Tab.~\ref{tab:ppm}. The results show that they have 
relatively similar performance, but PPM achieves the highest mF1 performance and has significantly lower latency than others.

\subsubsection{Effect of square supervision region}
For the training of the offset map and slope map predictions, we only supervise the square region of side length 
$2r+1$ centered on the groud-truth rail starting point. To verify the effectiveness of this square supervised region, we compare 
it to the commonly adopted supervised method~\cite{law2018cornernet} in which the training is only for keypoint locations. 
It is worth noting that the latter method is equivalent to ours when $r$ equals 0. As shown in Tab.~\ref{tab:radius}, the best performance 
can be achieved by setting $r$ to 2, and we improve the F1@50 metric by 0.51 compared to the latter method. 
This may be attributed to the fact that appropriately enlarging the supervision region can mitigate or even eliminate
the detrimental effects of starting point prediction bias.

\begin{table}[!htbp]
    \centering
	\caption{Ablation studies of different context aggregation ways}
    \label{tab:ppm}
    {
        \begin{tabular}{cccc}
            \hline
            Aggregation Ways & F1@50 & mF1 & Latency~(ms)\\ 
			\hline
            Self attention~\cite{Vaswani2017Attention} & \textbf{96.47} & 59.68 & 1.13  \\
            Efficient attention~\cite{shen2021Efficient} & 96.38 & 59.61 & 1.03 \\
            Strip pooling~\cite{hou2020spm} & 96.42 & 59.59 & 1.01 \\
			PPM & 96.43 & \textbf{59.79} & \textbf{0.52} \\
			\hline
        \end{tabular}
    }
\end{table}

\begin{table}[!htbp]
    \centering
	\caption{Ablation studies of different size of supervision region}
    \label{tab:radius}
    {
        \begin{tabular}{ccc}
            \hline
            Supervision region & F1@50 & mF1 \\ 
			\hline
            $r=0$ & 95.92 & 59.41 \\
			$r=1$ & 96.17 & 59.55 \\
            $r=2$ & \textbf{96.43} & \textbf{59.79} \\
            $r=3$ & 96.13 & 59.57 \\
			\hline
        \end{tabular}
    }
\end{table}

\subsection{Qualitative Results}
The qualitative results is showed in Fig.~\ref{fig:vis_results}. Our DALNet can generate a fitted anchor line 
for each rail or lane instance, even if it is occluded. And it achieves good rail or lane detection performance 
under different conditions.

\section{Conclusion}
In this paper, we propose a rail detection model DALNet based on dynamic anchor line along 
with DL-Rail, a scene-diverse urban rail detection dataset. In order to solve the problem that 
the predefined anchor lines are image agnostic, we introduce a dynamic anchor line generator to 
generate a fitted anchor line for each rail instance. We experimentally demonstrate that this 
dynamically constructed anchor line can be used as a better position reference for rail detection, and 
can significantly improve the performance of rail detection. 
In comparison with many fast lane detection methods, Our DALNet achieves state-of-the-art  
performance on the DL-Rail, Tusimple, and LLAMAS datasets, while still maintaining high efficiency.
\label{sec:conclusion}

\bibliographystyle{IEEEtran}
\bibliography{DALNet}

\end{document}